\documentclass[journal]{IEEEtran}

\usepackage{indentfirst}
\usepackage{enumerate}
\usepackage{multirow}
\usepackage{makecell}
\usepackage{cite}
\ifCLASSINFOpdf
   \usepackage[pdftex]{graphicx}
   \graphicspath{{../pdf/}{../jpeg/}}
   \DeclareGraphicsExtensions{.pdf,.jpeg,.png}
\else
   \usepackage[dvips]{graphicx}
   \graphicspath{{../eps/}}
   \DeclareGraphicsExtensions{.eps}
\fi

\usepackage[justification=centering]{caption}
\usepackage{amsmath}
\usepackage{algorithm}
\usepackage{algorithmic}
\usepackage{array}
\usepackage{subfig}
\usepackage{fixltx2e}
\usepackage{stfloats}
\usepackage{url}
\begin{document}
\title{Multi-objective Pointer Network\\ for Combinatorial Optimization}

\author{Le-yang~Gao,
        Rui~Wang,
        Chuang Liu,
        Zhao-hong~Jia% <-this % stops a space
\thanks{This work was supported in part by the National Natural Science Foundation of China (No. 71971002 and No. 71871076) and the National Science Fund for Outstanding Young Scholars (No. 62122093).}% <-this % stops a space
\thanks{L. Gao and Z. Jia (Corresponding author) are with the School of Computer Science and Technology, and C. Liu is with the School of Internet, Anhui University, Hefei 230039, PR China (e-mail: e20101008@stu.ahu.edu.cn; zhjia@mail.ustc.edu.cn; chuang@mail.ustc.edu.cn).}% <-this % stops a space
\thanks{Rui Wang is with the College of System Engineering, National University of Defense Technology,
Changsha 410073, PR China (ruiwangnudt@gmail.com).}
}

% line to be shifted left slightly. This is one of those LaTeX things. For
% instance, \textbf{A} \textbf{B} will typeset as A B not AB. To get
% AB then you have to do: \textbf{A}\textbf{B}

\markboth{ }%
{Shell \MakeLowercase{\textit{et al.}}: Bare Demo of IEEEtran.cls for IEEE Journals}



\maketitle

\begin{abstract}
Multi-objective combinatorial optimization problems (MOCOPs), one type of complex optimization problems, widely exist in various real applications. Although meta-heuristics have been successfully applied to address MOCOPs, the calculation time is often much longer. Recently, a number of deep reinforcement learning (DRL) methods have been proposed to generate approximate optimal solutions to the combinatorial optimization problems. However, the existing studies on DRL have seldom focused on MOCOPs. This study proposes a single-model deep reinforcement learning framework, called multi-objective Pointer Network (MOPN), where the input structure of PN is effectively improved so that the single PN is capable of solving MOCOPs. In addition, two training strategies, based on representative model and transfer learning, respectively, are proposed to further enhance the performance of MOPN in different application scenarios. Moreover, compared to classical meta-heuristics, MOPN only consumes much less time on forward propagation to obtain the Pareto front. Meanwhile, MOPN is insensitive to problem scale, meaning that a trained MOPN is able to address MOCOPs with different scales. To verify the performance of MOPN, extensive experiments are conducted on three multi-objective traveling salesman  problems, in comparison with one state-of-the-art model DRL-MOA and three classical multi-objective meta-heuristics. Experimental results demonstrate that the proposed model outperforms all the comparative methods with only 20\% to 40\% training time of DRL-MOA.
\end{abstract}


% \fi
\IEEEpeerreviewmaketitle

\section{Introduction}
% Here we have the typical use of a T for an initial drop letter
% and HIS in caps to complete the first word.
\IEEEPARstart{M}{ulti-objective} combinatorial optimization (MOCO)\cite{dominguez2021effective} is a class of mathematical optimization problems with discrete variables where multiple objectives are optimized simultaneously. MOCO problems (MOCOPs) can be formulated as follows\cite{kirlik2014new}:
\begin{equation}
\begin{aligned}
 &\mbox{Minimize}{~}F(X) = (f_1(X),f_2(X),\ldots,f_m(X))\\
 &\mbox{subject to}{~}X\in D
\end{aligned}
\label{eq:EQ1}
\end{equation}
where $X=(x_1,x_2,\ldots,x_n)$ represents an n-dimensional discrete candidate solution, $F(X)$  and $D\in R^n$ denote an M-dimensional objective space and an n-dimensional bounded discrete decision space, respectively. Due to several conflicting optimization objectives being considered simultaneously \cite{liu2020multi}, most MOCOPs are NP-hard\cite{cai2019decomposition}. Evaluating the quality of solutions to MOCOPs involves Pareto dominance relationship. A set of Pareto optimal solutions, also known as the non-dominated solution set, is obtained by solving MOCOP.

% that have no Pareto dominance relationship between each other

\indent
In recent years, a large number of MOCOPs have been extensively studied. The multi-objective traveling salesman problem (TSP), one of the typical MOCOPs, can be used to model a broad range of real-life applications, such as vehicle routing\cite{cheikhrouhou2021comprehensive}, manufacturing\cite{lokin1979procedures}, warehousing problem\cite{chisman1975clustered}, computer disk defragmentation\cite{zhang2018metaheuristics}, etc. Given a list of cities in TSP, an optimum feasible route that visits each city once and returns to the starting city is required. The TSP is one of the classical NP-hard problems. Therefore, multi-objective TSP (MOTSP) can be used to evaluate the capabilities of multi-objective solvers effectively. It is important to emphasize that in the TSP mentioned in this paper the costs between any two cities are the same in both directions.

\indent
Due to being able to obtain multiple Pareto-approximated optimal solutions in one single simulation execution, multi-objective evolutionary algorithms (MOEAs) are widely used to solve MOCOPs\cite{giagkiozis2015overview,zhou2011multiobjective}. Furthermore, individual recombination based on genetic operators and few hyperparameters make MOEA able to balance exploitation and exploration effectively. NSGAII\cite{deb2002fast} and SPEA2\cite{zitzler2001spea2}, two well-known MOEAs, have been widely employed in many real-world applications\cite{beirigo2016application,li2010application}. To improve the diversity of solutions when solving multi-objective optimization problems (MOOPs), evolutionary algorithms (EAs) based on decomposition have been proposed, such as NSGAIII\cite{deb2013evolutionary} and MOEA/D\cite{zhang2007moea}, to maintain sub-populations that are uniformly distributed in objective space, in order to ensure that their individuals are as uniformly close to the Pareto front as possible.

\indent
Over the past few years, artificial intelligence technology has been increasingly applied to address combinatorial optimization problems (COPs). Inspired by sequence-to-sequence model, Vinyals \textit{et al.}\cite{vinyals2015pointer} proposed Pointer Network (PN). The PN with Attention mechanism\cite{bahdanau2014neural} is trained to find approximate optimal solutions to TSP through supervised learning. However, high costs of constructing training samples with real optimal solutions hinder its application in large-scaled problems. Bello \textit{et al.}\cite{bello2016neural} adopted deep reinforcement learning (DRL) to train PN, with objective functions as feedback signal. At the same time, the critic network\cite{bello2016neural} was employed as the baseline to improve the stability of training, which not only reduces the consumption of building training dataset, but also surpasses the original PN on addressing TSP and 0-1 knapsack problem. Nazari \textit{et al.}\cite{nazari2018reinforcement} applied PN to vehicle routing problem (VRP) with dynamic properties and replaced the recurrent neural network (RNN) with a one-dimensional convolutional layer to receive the input. This model decreases the training time by 60\% without weakening its optimization capability. At this point, the PN using DRL has demonstrated the following significant advantages over the other combinatorial optimization methods. 
\begin{itemize}
\item Strong optimization capability, without requiring manually designed optimization strategies and heuristics.
\item Fast optimization, that is, the trained model solves the problem through forward propagation without iterative process similar to EAs.
\item Insensitive to problem scale, that is, the trained model can be applied to similar problems with different scales.
\end{itemize}

\indent
The use of DRL and PN to solve MOCOPs is still in the early stages. Li \textit{et al.}\cite{li2020deep} proposed a multi-objective deep reinforcement learning framework, called DRL-MOA, which divides the MOCOP into several sub-problems. These sub-problems are solved separately using the corresponding number of sub-PNs, which are trained collaboratively with a neighbourhood-based parameter transfer approach. According to the experimental results, DRL-MOA shows better optimization performance in three types of MOTSP problems. It is obvious that DRL-MOA requires training a large number of PNs to ensure that the complete Pareto front is obtained, which results in expensive training consumption.

\indent
To further improve the utility of DRL models in the field of MOCO, this paper proposes a single-model multi-objective Pointer Network (MOPN) using DRL. The main contributions of this study are as follows:
\begin{itemize}
\item The input structure of the encoder of PN is improved with the incorporated objective weight, so that the single PN is also capable of solving MOCOPs, which reduces the training time of MOPN significantly compared to DRL-MOA.
\item Two effective training strategies are proposed to further improve the training efficiency and optimization capability of the MOPN on MOCOPs.
\item A valuable conclusion is obtained from experimental results. Instead of a larger-scaled training dataset, the training dataset with a suitable problem scale can make MOPN easier to obtain good training effectiveness and stronge generalization ability. 
\end{itemize}

\indent
The structure of the paper is organized as follows: Section \uppercase\expandafter{\romannumeral2} introduces the proposed MOPN in detail. The design of the comparative experiment is presented in Section \uppercase\expandafter{\romannumeral3}. The experimental results and discussions are described in Section \uppercase\expandafter{\romannumeral4}. Finally, the conclusion of this study and future research directions are given in Section \uppercase\expandafter{\romannumeral5}.

\section{Multi-Objective Pointer Network for Combinatorial Optimization}
In this section, the proposed single-model MOPN to address MOCOP is introduced in detail. Firstly, the mathematical definition of MOTSP studied in this paper is formulated. Then, the framework of the proposed model is described, with an emphasis on the improvement of the input structure to the encoder. Finally, the training method and two proposed training strategies are presented.

\subsection{Formulation of MOTSP}
The MOTSP with $M$ optimization objectives can be defined as follows: It is assumed that there is a set of $n$ cities $X$, denoted by $X=\{x^1,x^2,\ldots,x^n\}$, where moving from city $i$ to city $j$ generates $M$ kinds of costs, denoted by $(c_{ij}^1,c_{ij}^2,\ldots,c_{ij}^M)$. Tour $Y$ that visits each city exactly once and returns to the starting point needs to be found while minimizing $M$ cost functions simultaneously. The $m$-th cost function for tour $Y$ can be formulated as follows:
\begin{equation}
\begin{aligned}
 &C_m(Y)=\sum_{i = 1}^{n-1}c_{y_iy_{i+1}}^m+c_{y_ny_1}^m, m=1,2,\ldots,M
\end{aligned}
\label{eq:EQ2}
\end{equation}
where $y_i$ denotes the $i$-th visited city of $Y$.

\indent
Suppose there is a set of weight vectors $w^1,w^2,\ldots$ that uniformly distributed in the objective space. Take the bi-objective optimization problem as an example. A set of 100 vectors, $(1,0),(0.99,0.01),\ldots,(0,1)$, is selected as weight vectors. Each weight vector $w^K=\{w_1^K,w_2^K,\ldots,w_M^K\}$ represents a user preference for the MOCOP, and the elements in the weight vector represent the priorities of objectives, where larger value means higher priority. It is notable that $w_1^K+w_2^K+\ldots+w_M^K=1$. Accordingly, the aggregated multi-objective cost function of tour $Y$ with user preference $w^K$ is formulated as follows:
\begin{equation}
\begin{aligned}
 &C(Y)=\sum_{m=1}^{M}{w_m^Kc_m(Y)}
\end{aligned}
\label{eq:EQ3}
\end{equation}

%\indent
%The aim of this work is to obtain the approximate optimal solution to the MOTSP under any preferences through the proposed single-model MOPN.

\subsection{Framework of MOPN}
In this part, a generic model MOPN is designed for MOCOP, with a given set of inputs $X=\{x^1,x^2,\ldots,x^n\}$. For MOTSP, each element in $X$ represents the set of features of the corresponding city, i.e., $x^i=\{x_1^i,x_2^i,\ldots\}$, where $x_j^i$ represents the $j$-th feature of the $i$-th city. And the number of city features is determined according to the definition of the problem. Moreover, the output of the proposed model is defined as $Y=(y_1,y_2,\ldots,y_n)$, which can represent the requested tour in MOTSP, where $y_t$ denotes the index of the city chosen to be visited at step $t$. Accordingly, the probability of selecting output $Y$ with input $X$ through the probability chain rule is given as follows\cite{gu2020pointer}:
\begin{equation}
\begin{aligned}
 &P(Y|X)=\prod_{t=1}^{n}{P(y_{t+1}|y_1,y_2,\ldots,y_t,X_t)}
\end{aligned}
\label{eq:EQ4}
\end{equation}
where $X_t$, representing the set of all available cities at step $t$, is updated after each city is visited. $P(y_{t+1}|y_1,y_2,\ldots,y_t,X_t)$ denotes the probability of selecting $y_{t+1}$ as the next city to be visited after $y_t$, based on the visited cities being $y_1,y_2,\ldots,y_t$. The process of determining the next city to be visited according to conditional probabilities is referred to as a step. Therefore, the construction of a feasible tour can be obtained througn performing this step iteratively. As a result, the process of mapping from input $X$ to output $Y$ can be modelled via PN.

\begin{figure}
\centering
\includegraphics[width=1\linewidth]{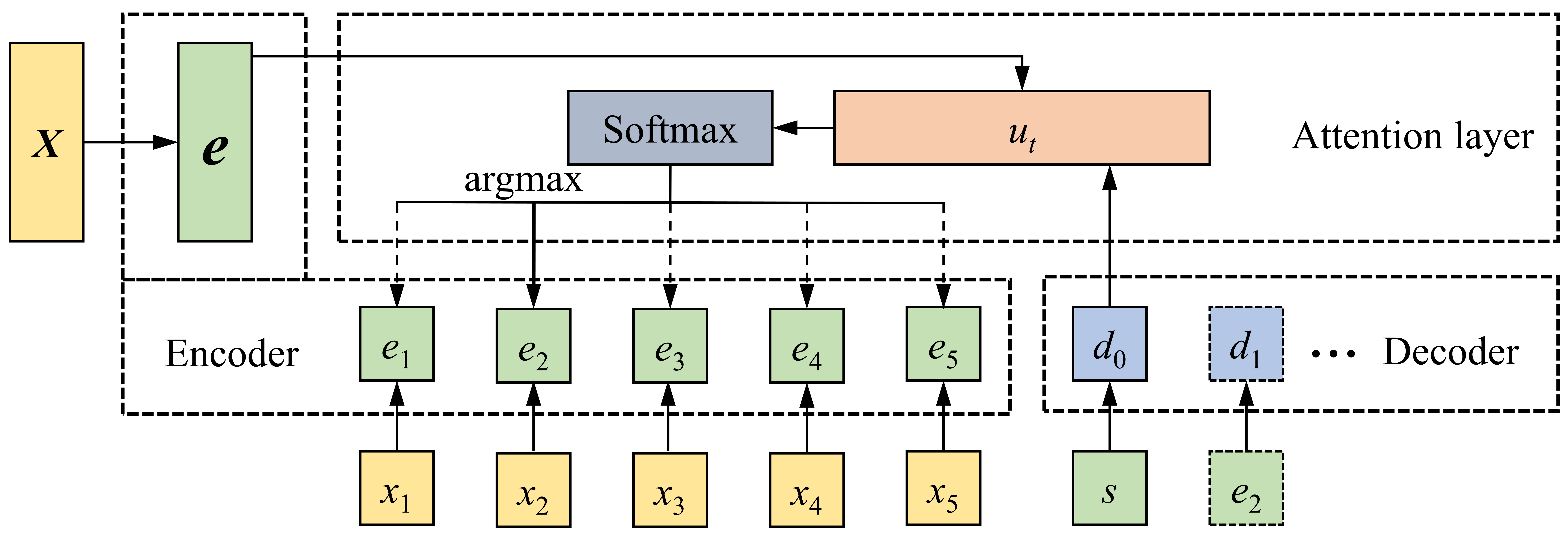}
\caption{Structure of the MOPN}
\label{fig:FIG1}
\end{figure}

\indent
In this paper, with the improved input structure of the encoder and the proposed training strategies, a novel PN model, called MOPN, is proposed to solve MOCOPs. Fig.\ref{fig:FIG1} shows the structure of the MOPN. In the following, the three modules of the model, that is, the encoder, the decoder, and the attention layer, are described in detail, respectively.

\subsubsection{Encoder}
\indent
 
\indent
The encoder can recognize the input sequence and store the features of cities in a vector with fixed size. Classical PN generally employs an RNN\cite{cho2014learning} as the encoder, which can aggregate both the input information and the input order, such as the words in a sentence and the order of the words. However, the inputs to many COPs are mutually independent entities with no temporal relationship, such as cities in TSP and VRP\cite{zajac2021objectives}, items in knapsack problem\cite{zhou2021hybrid}, and jobs in scheduling problem\cite{tomazella2020comprehensive}, etc. For solving such problems, the RNN in encoder is no longer appropriate. Therefore, based on\cite{nazari2018reinforcement}, an one-dimensional convolutional layer is employed, which not only maps the input to a high-dimensional vector, but also has a smaller computational complexity than RNN.

\indent
In addition, in order to satisfy the needs for solving MOCOPs, the building rules related to the input vector of the encoder are presented as follows:
\begin{enumerate}[a)]
\item The feature vector of city and weight vector of objectives are stitched together to form a input vector. The order of the objectives corresponding to the elements in the feature vector should be the same as that of weight vector.
\item In the feature vector, the feature dimension associated with each objective is kept consistent. If the feature dimension of one objective is less than others, it is necessary to pad the features according to the corresponding dimension. The padded features are set to 1.
\end{enumerate}

\begin{figure}
\centering
\includegraphics[width=1\linewidth]{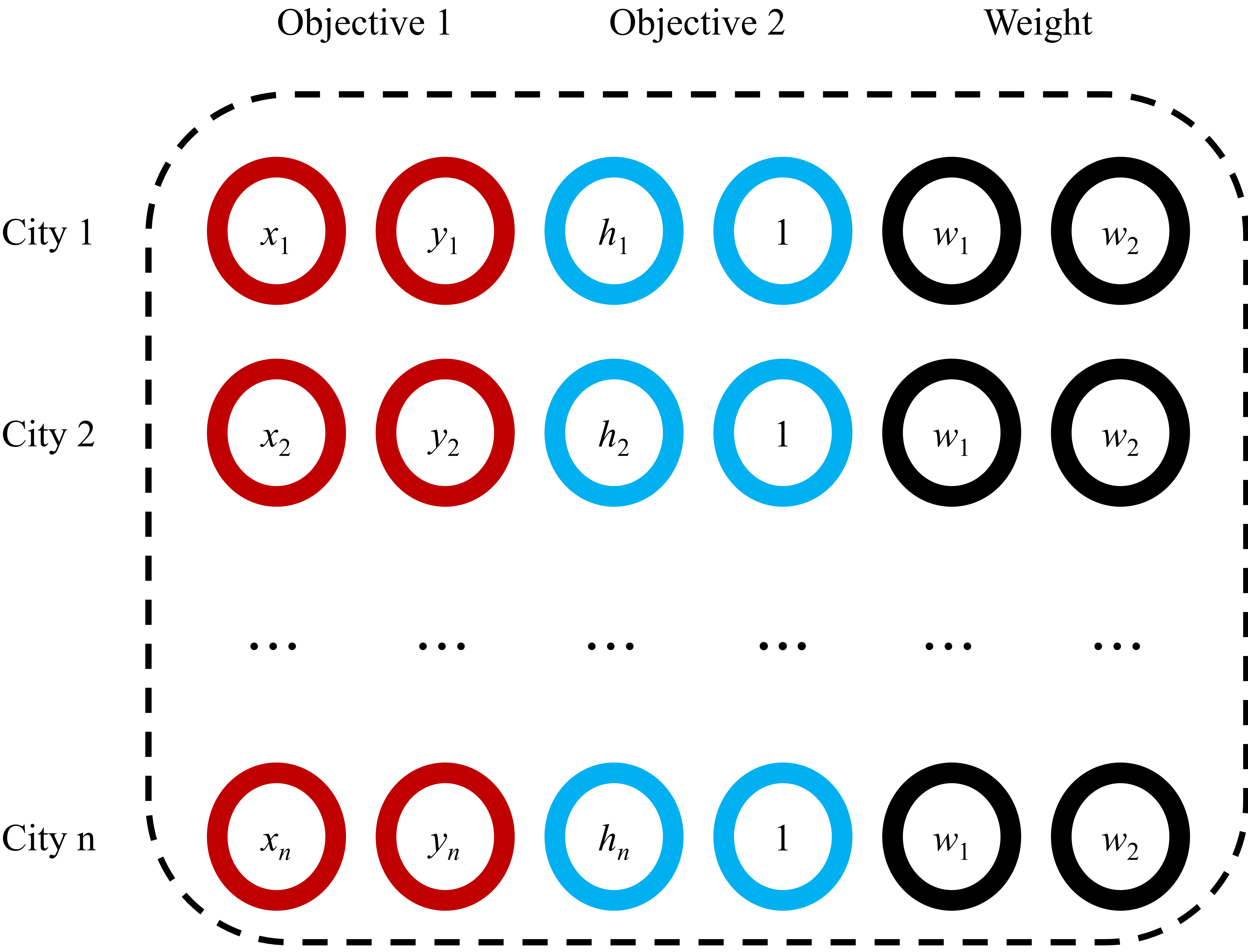}
\caption{Input vector of the encoder for an example MOTSP}
\label{fig:FIG2}
\end{figure}

\indent
By the building rules, the order and position of elements in the input vector are given practical meaning, which increase the amount of information that the input vector provides to the model. Meanwhile, the weights of objectives are also included into the input vector, which ensures that the information sent to all subsequent modules contains the objective weights. To explain the building rules of input vector specifically, an example for a two-objective MOTSP is provided. Suppose that the first objective is to minimize the tour length determined by the two-dimensional coordinates $(x_i,y_i)$ of one city, and the second objective is to minimize the sum of altitude differences between neighbouring cities, where the altitude of city $i$ is represented by $h_i$. $w_1$ and $w_2$ denote the weights of the two objectives, respectively. Therefore, the input vector of the encoder for this problem can be illustrated in Fig.\ref{fig:FIG2}.

\indent
As the feature of the second objective is only one-dimensional, according to the second rule, the fourth column in Fig.\ref{fig:FIG2} is added to be consistent with the feature dimensions of the other objectives. Furthermore, as can be seen from Fig.\ref{fig:FIG2}, the objective weights are present at the end of input vector for each city.

\indent
Through the encoder, the feature vector of each city is transformed into a vector with length $l$. These vectors, called the node embedding, are used as the input to the decoder. The node embedding of all cities can be merged into a matrix of $n*l$, called context embedding, which is a part of the input to the attention layer. It is worth noting that $l$ represents the size of the one-dimensional convolutional layer of the encoder and $n$ denotes the number of cities.

\subsubsection{Decoder}
\indent
 
\indent
RNN is the main component of the decoder, which stores all the information of node embedding that has been input into the decoder. By combining this historical information with the node embedding of the last visited city, the decoder can output a decoding vector containing information about all the cities having been visited so far. This decoding vector, that is, the hidden layer state $d_t$ of the RNN, is used as the input to the attention layer in step $t$.

\indent
In order to minimize the consumption of training model, an RNN model of gated recurrent unit (GRU)\cite{cho2014properties}, with similar performance of the LSTM\cite{hochreiter1997long} but with fewer parameters, is adopted in this study.

\subsubsection{Attention layer}
\indent
 
\indent
In each step, the attention layer receives the context embedding $e$ and the decoding vector $d_t$ from the encoder and decoder, respectively, and then calculates the correlation of each city with the next visited city. The city with the highest correlation is most likely to be selected as the next city. The process is formulated as follows\cite{nazari2018reinforcement}:
\begin{equation}
\begin{aligned}
 &u_t^i=v_a^T\mbox{tanh}(W_a[e_i;d_t]),i=1,2,\ldots,n
\end{aligned}
\label{eq:EQ5}
\end{equation}
\begin{equation}
\begin{aligned}
 &a_t=\mbox{softmax}(u_t)
\end{aligned}
\label{eq:EQ6}
\end{equation}
\begin{equation}
\begin{aligned}
 &b_t=a_te^T
\end{aligned}
\label{eq:EQ7}
\end{equation}
\begin{equation}
\begin{aligned}
 &\tilde{u}_t^i=v_b^T\mbox{tanh}(W_b[e_i;b_t]),i=1,2,\ldots,n
\end{aligned}
\label{eq:EQ8}
\end{equation}
\begin{equation}
\begin{aligned}
 &P(y_{t+1}|y_1,y_2,\ldots,y_t,X_t)=\mbox{softmax}(\tilde{u}_t)
\end{aligned}
\label{eq:EQ9}
\end{equation}
where ";" means the concatenation of two vectors, $v_a$, $v_c$, $W_a$, and $W_c$ are all learnable parameters, $a_t$ and $b_t$ denote the "attention" mask over the inputs and the context vector in step $t$, respectively, and $\tilde{u}_t^i$ represents the likelihood that city $i$ is chosen as the city to be visited in the next step. The softmax operator is used to normalize $u_t$ and $\tilde{u}_t$. Ultimately, the probability of each city being visited at step $t+1$ is obtained.

\indent
Assume that the matrix recording the objective features of all cities is a root instance (RIns) of MOTSP. By a set of objective vectors uniformly distributed in objective space, a root instance can be extended into many leaf instances (LIns) with different preferences. Each LIns is denoted by a matrix similar to Fig.\ref{fig:FIG2}. Based on the three main modules of MOPN, as shown in Fig.\ref{fig:FIG1}, the forward propagation process is described in Algorithm \ref{alg:ALG1}.

\begin{algorithm}[t]
\raggedright
\caption{Forward propagation process of MOPN}
\label{alg:ALG1}
{\bf Input:} LIns $X$, number of cities $n$, zero vector $s$ of length $l$\\
{\bf Output:} tour $Y$ with corresponding preference\\
\begin{algorithmic}[1]
\STATE Initialize a new empty tour $Y$
\STATE $e\gets$ Encoder($X$)
\FOR {$t\gets$$ 1,2,\dots,n$}
    \IF{$t=1$}
        \STATE $d_t\gets$ Decoder($s$)
    \ELSE
        \STATE $d_t\gets$ Decoder($e_i$)
    \ENDIF
    \STATE $\tilde{u}_t\gets$ Attention layer($e, d_t$)
    \STATE $P\gets$ softmax$(\tilde{u}_t)$
    \STATE Greedily select city $i$ with the highest probability in $P$
    \STATE $Y\gets Y\cup\{i\}$
\ENDFOR
\end{algorithmic}
\end{algorithm}

\indent
During training, instead of greedily selecting, the MOPN selects the next city through sampling from the probability distribution.

\indent
After Algorithm \ref{alg:ALG1} is executed with a LIns set of one RIns as the input, a set of solutions converging with different objective weights is obtained. The Pareto non-dominated solutions among the obtained solutions are selected to form the Pareto front of this RIns.

%More details of the operator to obtain the Pareto front are provided in Section \uppercase\expandafter{\romannumeral3}.B.

\subsection{Training method for MOPN}
To train the MOPN, the well-known Actor-Critic policy gradient method is adopted\cite{peters2008natural}. The policy gradient method can be used to iteratively train all trainable parameters of the encoder, decoder and attention layer based on the gradient estimate of the expected reward.

\indent
In this paper, the policy gradient method consists of an actor network and a critic network, whose parameters are $\theta$ and $\delta$, respectively. In brief, the main part of the actor network, the proposed MOPN, can map the given LIns of MOCOP to a probability distribution of solutions. Then, the actor network obtains an expected solution by sampling from this probability distribution and computes its objective function $C$ as the reward according to Eq.(\ref{eq:EQ3}). The critic network estimates the expected reward $Z(X;\delta)$ of the solution obtained by the actor network based on information about the given LIns. In this paper, the proposed MOPN is trained using the policy gradient method based on batch training. Therefore, the average gradients of the actor and the critic networks for $N$ LIns are formulated as follows\cite{deudon2018learning}:
\begin{equation}
\begin{aligned}
 &d\theta=\frac{1}{N}\sum_{k=1}^N(C^k-Z(X^k;\delta))\bigtriangledown_{\theta}\log{}P(Y^k|X^k)
\end{aligned}
\label{eq:EQ10}
\end{equation}
\begin{equation}
\begin{aligned}
 &d\delta=\frac{1}{N}\sum_{k=1}^N\bigtriangledown_{\delta}(C^k-Z(X^k;\delta))^2
\end{aligned}
\label{eq:EQ11}
\end{equation}
where $C$ and $Z(X^k;\delta)$ represent the expected rewards obtained through the actor and the critic networks on LIns $k$, respectively.

\indent
In addition, the policy gradient is an unsupervised training method, meaning that the optimal objective values of LIns is not required for training. Since the training dataset only consists of random LIns, the generation cost of dataset is almost negligible.

\subsection{Proposed training strategy}
Considering the high complexity of MOCOPs and the low cost of dataset generation, the training dataset of MOPN needs to contain a large number of LIns to improve the performance, which also leads to high time consumption for training. In order to balance between model performance and training time, as well as being applied to different application scenarios, the following three training strategies are proposed.

\subsubsection{General training strategy (GTS)}
\indent
 
\indent
In this strategy, the model to solve MOTSP with a given scale is obtained by training on the dataset with the corresponding scale. The disadvantages of this strategy are mainly two aspects. On the one hand, the training of MOPN will be time-consuming when the problem scale is large. On the other hand, this strategy makes the MOPN no longer insensitive to problem scale.

\subsubsection{Training strategy based on representative model (TS-RM)}
\indent
 
\indent
In this strategy, a representative model (RM) is obtained by training on the dataset with an appropriate problem scale. RM is expected to perform well on problems with most scales. In order to find the RM that meets the above requirements, the problem scale of the training dataset needs to be determined through a process of parameter adjustment. This strategy reduces the training time of MOPN while ensuring that the obtained model is insensitive to the problem scale.

\subsubsection{Training strategy based on transfer learning (TS-TL)}
\indent
 
\indent
Considering that the transfer learning is applicable to training PN\cite{li2020deep}, a training strategy based on transfer learning (TS-TL) is proposed to enhance the performance of MOPN. In this strategy, RM is adopted as an initial model for transfer learning. After the initial model is trained on the dataset with other scale for a short period of time, it will obtain better performance for problems of the corresponding scale. This strategy is able to balance between the time consumption of training and the capability of MOPN. However, the obtained model is sensitive to problem scale.

\indent
The details of the parameter adjustment about these strategies is specified in Section \uppercase\expandafter{\romannumeral3}.C.

\section{Experiment setup}
In order to evaluate the performance of the proposed MOPN comprehensively, all-round comparative experiments are designed involving MOTSPs with three types of objective combinations in this section. The comparative methods include a state-of-the-art multi-objective optimization model DRL-MOA and three classical multi-objective meta-heuristics, that is, MOEA/D, NSGAII and NSGAIII. All training of the models is conducted on a single GTX 2080Ti GPU to compare their training efficiency. And all models and meta-heuristics are run on the Intel Core i7-10875H CPU with 16GB of RAM. The proposed models are implemented in python and are publicly available on GitHub\footnote{https://github.com/gaoly/MOPN} to reproduce the experimental results and facilitate further research. The DRL-MOA is also publicly available on GitHub\footnote{https://github.com/kevin031060/RL\_TSP\_4static}. And the comparative meta-heuristics are implemented by calling the Pymoo Library\cite{blank2020pymoo}. 

\subsection{Introduction to testing problem}
In this paper, two objectives of MOTSP are adopted in testing problem.

\indent
One of the objectives, tour length, is most widely considered in MOTSPs. When calculating this objective function using Eq.(\ref{eq:EQ2}), the cost of travel between two cities is defined as their Euclidean distance. Minimizing this objective can effectively reduce the fuel cost of travel. The MOTSP instances with minimizing tour length contain the coordinates $(x_i,y_i)$ of each city.

\indent
The second objective is the sum of the altitude differences between cities in a tour. It is usually used to reflect the smoothness of a travel route. Minimizing this objective can reduce fuel costs and increase travel comfort simultaneously. This objective is also calculated according to Eq.(\ref{eq:EQ2}), where $c_{ij}$ represents the absolute value of the altitude differences between two cities. The MOTSP instances with this objective contain the altitude $h_i$ of each city.

\indent
In experiments, the testing problems involving one and two types of optimization objectives are encoded as T1 and T2, respectively. O2 and O3 represent the testing problems have two and three optimization objectives, respectively. The codes of the testing problems are shown in TABLE \ref{tab:TAB1}.
\begin{table}[!htbp]
\centering
\caption{Testing problems}
\label{tab:TAB1}
\resizebox{\linewidth}{!}{
  \begin{tabular}{cccc}  \hline
    \multicolumn{1}{l}{Problem Code} & Obj1 & Obj2 & Obj3        \\ \hline
        T1O2  & Tour Length & Tour Length & -        \\
        T2O2  & Tour Length & Altitude Difference & -        \\
        T2O3  & Tour Length & Tour Length & Altitude Difference        \\  \hline
  \end{tabular}
}
\end{table}

\indent
The two objectives of problem T1O2 are identical, complex and in conflict with each other. This kind of problems is suitable for verifying the exploitation and exploration capabilities of the proposed model. The objectives of problem T2O2 are tour length and altitude difference, respectively. This type of problem demonstrates the ability of the proposed model to optimize the problem with different objectives. In order to evaluate the performance of the model for optimizing problem with more than two objectives, problem 'T2O3' is also proposed.

\indent
It is worth mentioning that problems T1O2 and T2O3, with two same objectives of tour lengths, are presented in TABLE \ref{tab:TAB1}. In this type of RIns, each city has two coordinates, which exist in different coordinate systems, respectively. Furthermore, the tour lengths in the two coordinate systems are calculated independently.

\indent
Considering different problem scales, each type of the testing problems is extended into 40-, 70-, 100-, 150- and 200-city MOTSPs.

Based on three types of testing problems and five problem scales, fifteen levels of testing problems are generated. For example, T1O2S100 represents problem T1O2 with 100-city. The training dataset of each level contains 500000 LIns according to \cite{li2020deep}. All LIns are generated according to uniform distribution within [0,1]. Their weight vectors are also normalized. Different from the training dataset, the testing dataset contains RIns, instead of LIns. The DRL model needs to obtain the Pareto front by combining RIns with a set of weight vectors, as described in Section \uppercase\expandafter{\romannumeral2}.B. Moreover, the meta-heuristics can get the Pareto front directly by RIns. Considering that the meta-heuristics require a large number of iterations and take long computation time to solve MOTSPs, the number of RIns in the testing dataset is set to 20 to control the time consumption of experiment.

\indent
Besides the generated testing datasets, the widely used RIns kroAB100, kroAB150 and kroAB200 from TSPLIB\cite{reinelt1991tsplib}, corresponding to the 100-, 150- and 200-city T1O2 problems respectively, are also adopted. Note that the city features of these RIns also need to be normalized.

\subsection{Introduction to evaluation indicators}
In this paper, four evaluation indicators are selected to evaluate the performance of the comparative methods.

\subsubsection{Hypervolume}
\indent
 
\indent
The hypervolume\cite{while2006faster}, denoted by HV, describes the size of the region dominated by the non-dominated solution set. The larger the HV value, the closer the non-dominated solution set is to the real Pareto front, indicating better convergence and diversity of this set. The calculation of the HV relies on the selection of the reference point. In this paper, the maximum of each objective in the output solution sets obtained by all comparative methods is dynamically selected as the corresponding component of the reference point.

\subsubsection{Spacing}
\indent
 
\indent
Spacing\cite{deb2002fast}, denoted by SPC, measures the uniformity of the propagation of the set of non-dominated solutions. The calculation of SPC involves the extreme point, the Euclidean distance $D_i$ between consecutive non-dominated solutions and its average value $\overline{D}$. The extreme points represent the non-dominated solutions that are optimal in each objective direction, respectively, and are selected in the same way as the reference point related to HV. The SPC for the bi-objective problem is formulated in Eq.(\ref{eq:EQ12}).
\begin{equation}
\begin{aligned}
 &\mbox{SPC}=\frac{D_f+D_l+\sum_{i=1}^{N-1}{|D_i-\overline{D}|}}{D_f+D_l+(N-1)\overline{D}}
\end{aligned}
\label{eq:EQ12}
\end{equation}
where $D_f$ and $D_l$ represent the distances from the two extreme points to their nearest point of the non-dominated solution set, respectively. The smaller the value of SPC, the more uniform the distribution of the solution set in the objective space. To be consistent with the definition of SPC, the three-objective problems is partitioned into three bi-objective problems, whose objectives are defined as every two of the three objectives. The non-dominated solutions to the three-objective problem are projected into the coordinate systems corresponding to the three bi-objective problems, and are selected according to Pareto dominance relationship to obtain Pareto fronts in the three coordinate systems. The average SPC value of these Pareto fronts is proposed to evaluate the uniformity in three-objective problems.

\subsubsection{Training time}
\indent
 
\indent
The training time denotes the time taken by the model from initialization to completion of training. Since the meta-heuristics do not require training, MOEA/D, NSGAII and NSGAIII are not involved in the comparison of this indicator.

\subsubsection{Running time}
\indent
 
\indent
The running time represents not only the time taken by the model to obtain the Pareto front through forwarding propagation, but also the time spent by the meta-heuristic to complete its iterations. Both training time and running time reflect the practicality and usability of the comparative methods.

\subsection{Setting of parameters}
The parameters of the proposed MOPN and all comparative methods are determined according to their original paper or preliminary experiments. 

\subsubsection{Generation of reference vectors}
\indent
 
\indent
In this paper, the set of weight vectors, that is uniformly distributed in the objective space, is used in the execution of MOPN, DRL-MOA, MOEA/D and NSGAIII. These vectors, referred to as reference vectors, are generated by the simplex-lattice design method\cite{cornell2011experiments}. The number of reference vectors about the two-objective and three-objective problems are set to 100 and 105, respectively. And the generation parameters of the simplex-lattice design method for the two-objective and three-objective problems are set to $(100,0)$ and $(13,0)$, respectively.

\subsubsection{Setting of network parameters for MOPN and DRL-MOA}
\indent
 
\indent
The network parameters of MOPN are set to be similar to those of the models in \cite{li2020deep} that are shown in TABLE \ref{tab:TAB2}.
\begin{table*}[!htbp]
\centering
\caption{Network parameters of the MOPN}
\label{tab:TAB2}
\setlength\tabcolsep{10pt}%调列距
  \begin{tabular}{ccc}  \hline
    \multicolumn{3}{c}{Actor Network}       \\ \hline
        Module  & Type & \multicolumn{1}{l}{Parameter}          \\ \hline
        Encoder  & 1D-Conv & \multicolumn{1}{l}{$D_{input}=D_{problem}$, $D_{output}=128$, kernel size=1, stride=1}\\
        Decoder  & GRU & \multicolumn{1}{l}{$D_{input}=128$, $D_{output}=128$, hidden size=128, number of layer=1}        \\
        Attention  & - & \multicolumn{1}{l}{No hyper parameters}       \\  \hline
    \multicolumn{3}{c}{Critic Network}       \\ \hline
        First layer  & 1D-Conv & \multicolumn{1}{l}{$D_{input}=D_{problem}$, $D_{output}=128$, kernel size=1, stride=1}\\
        second layer  & 1D-Conv & \multicolumn{1}{l}{$D_{input}=128$, $D_{output}=20$, kernel size=1, stride=1}\\
        Third layer  & 1D-Conv & \multicolumn{1}{l}{$D_{input}=20$, $D_{output}=20$, kernel size=1, stride=1}\\
        Forth layer  & 1D-Conv & \multicolumn{1}{l}{$D_{input}=20$, $D_{output}=1$, kernel size=1, stride=1}\\  \hline
  \end{tabular}
%\vspace{0.1in}
\end{table*}

\indent
$D_{input}$ and $D_{output}$ in TABLE \ref{tab:TAB2} represent the input dimension and output dimension of the network, respectively. $D_{problem}$ denotes the input dimension of the problem. "1D-Conv" means the one-dimensional convolutional layer.
"Kernel size" and "stride" are important parameters for the convolutional calculation. In the decoder, a one-layer GRU network with the hidden size of 128 is employed. As can be seen from TABLE \ref{tab:TAB2}, the critic network consists of four one-dimensional convolutional layers connected in a sequence. The final output of critic network is a one-dimensional evaluation value.

\indent
All the learnable parameters of the proposed model are initialized through the Xavier initialization method\cite{glorot2010understanding}. The Adam optimiser\cite{kingma2014adam} is used to optimize them, with learning rate of 0.0001 and batch size of 200.

\indent
The parameters of each sub-model for DRL-MOA are also based on\cite{li2020deep}. It is worth noting that the epoch number for training each sub-model of DRL-MOA is set to 5. This means that the total number of epochs of DRL-MOA is five times the number of sub-models. The numbers of sub-models of DRL-MOA are set to 100 and 105 for the two- and three-objective problems, respectively. Therefore, the corresponding total numbers of epochs for training are 500 and 525, respectively.

\subsubsection{Setting of parameters for the training strategies}
\indent
 
\indent
In Section \uppercase\expandafter{\romannumeral2}.D, three training strategies are proposed to balance between model performance and training time. In these strategies, the problem scales of training datasets for RM need to be determined by preliminary experiments.

\indent
The selected MOTSP in the preliminary experiments is T1O2, due to sufficient complexity and shorter training time. In the preliminary experiments, six proposed models are trained on the datasets with problem scales of 10, 20, 30, 40, 50 and 100, respectively. The epoch number of training is set to 100. In addition, the problem scales of testing datasets are set to 10, 100 and 200 to evaluate the model performance for problems with different scales. Each model obtains the sets of non-dominated solutions for all 20 RIns in each testing dataset so that the average of their HV values is calculated. 

\begin{figure*}
  \centering
  \subfloat[Box plots on T1O2 MOTSP with 10-city]{
        \includegraphics[width=0.3\textwidth]{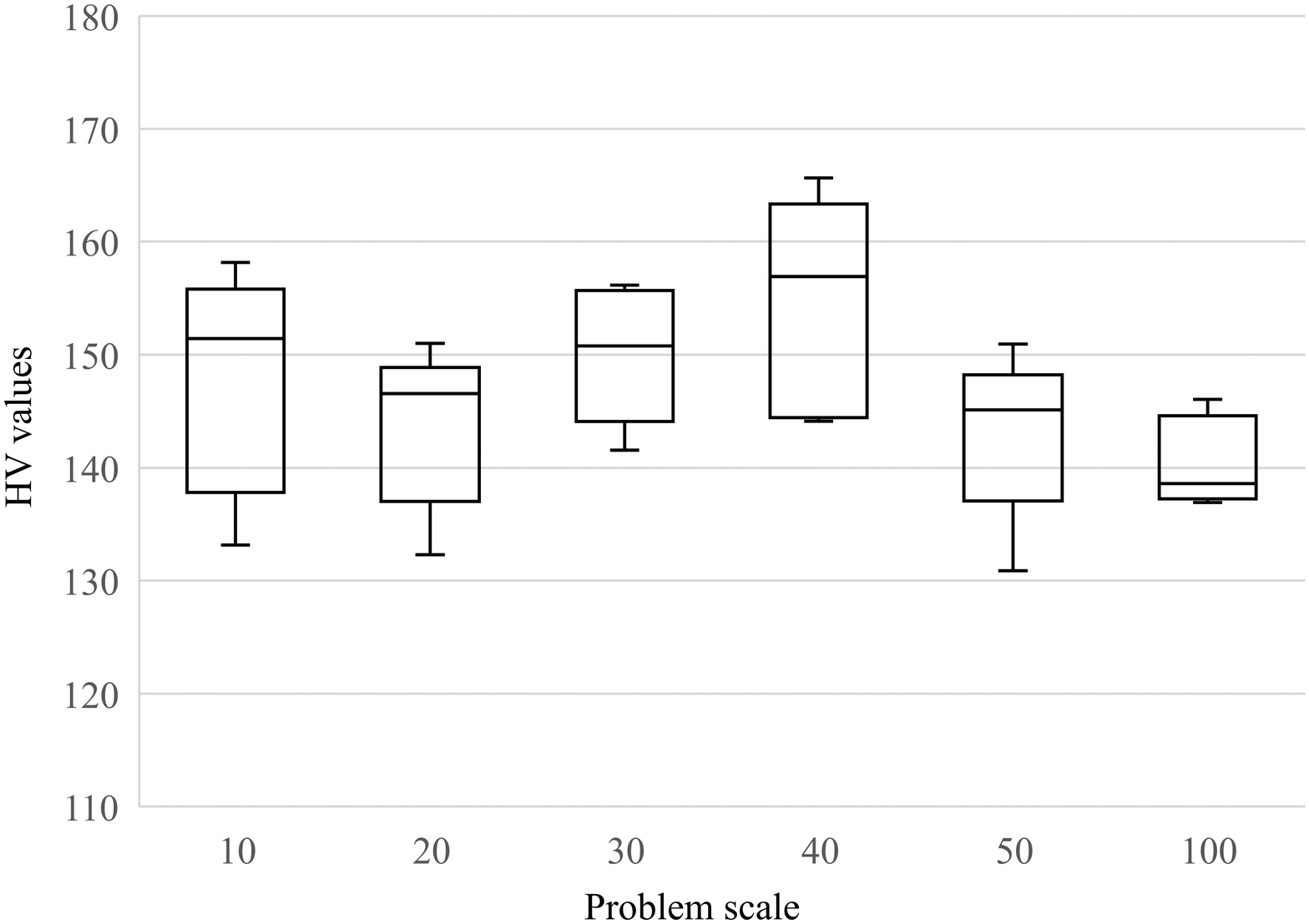}
         \label{fig:FIG3-1}}
  \quad
  \subfloat[Box plots on T1O2 MOTSP with 100-city]{
        \includegraphics[width=0.3\textwidth]{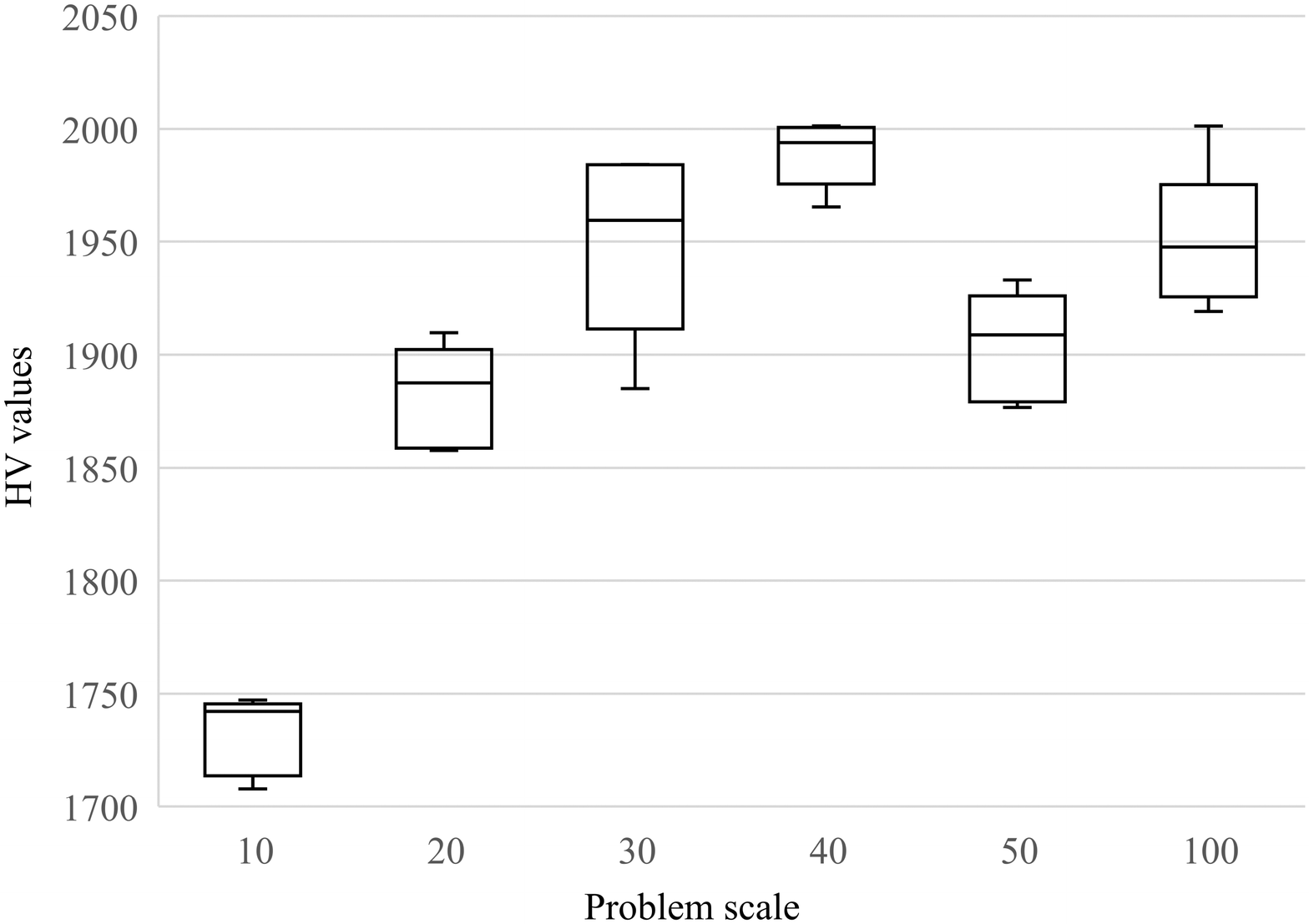}
        \label{fig:FIG3-2}}
  \quad
  \subfloat[Box plots on T1O2 MOTSP with 200-city]{
        \includegraphics[width=0.3\textwidth]{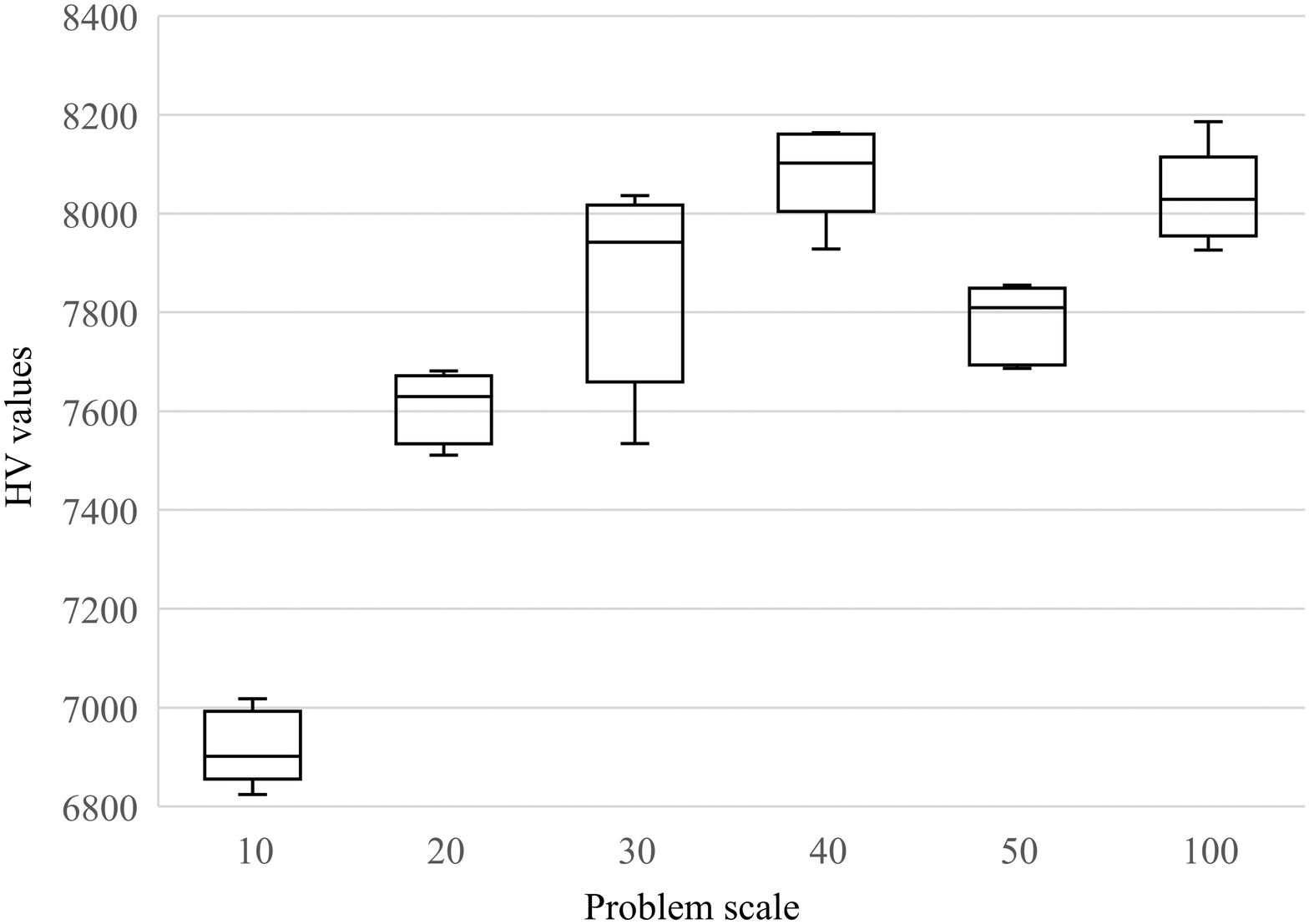}
        \label{fig:FIG3-3}}
  \caption{Box plots of the results of the preliminary experiment}
  \label{fig:FIG3}
\end{figure*}

\indent
The three box plots of the results of the preliminary experiments are shown in Fig.\ref{fig:FIG3}. According to Fig.\ref{fig:FIG3-1}, on the MOTSP with 10-city, models trained on 40-city show better performance than those trained on the dataset with the same scale as that of the MOTSP. As can be seen from Fig.\ref{fig:FIG3-2} and Fig.\ref{fig:FIG3-3}, the models trained on 10-city, 20-city and 50-city perform poorly on the testing datasets with large scales. Moreover, the models trained on 40-city show the best optimization capability and robustness for the problems in Fig.\ref{fig:FIG3-2} and Fig.\ref{fig:FIG3-3}. Note that none of the models trained on 10-city and 100-city obtain the desired performance in the corresponding testing datasets. It be concluded that: 1) the LIns with smaller scale do not have complete knowledge to solve MOTSP; 2) Secondly, the LIns with larger scale may cause MOPN to fall into local optimum during training. Therefore, the results of the preliminary experiments demonstrate that the GTS is weaker than TS-RM overall. The subsequent experiments do not include the GTS in Section \uppercase\expandafter{\romannumeral2}.D. Finally, the dataset with problem scale of 40 is selected as the training dataset for RMs.

\indent
According to the proposed TS-TL, RM should be trained on the dataset with the corresponding problem scale to improve performance, called the transfer learning phase. In this study, the scale of the training dataset used for this phase is set to 100 to control time consumption. To ensure that RM converges as much as possible, the number of training epochs is set to 100. Considering the high time-consumption training on 100-city, the number of training epochs in the transfer learning phase is set to 30. 

\subsubsection{Setting of parameters for the meta-heuristics}
\indent
 
\indent
The classical multi-objective meta-heuristics, MOEA/D, NSGAII and NSGAIII, are selected as the comparative algorithms. MOEA/D is involved in comparative experiments for all problems, and NSGAII and NSGAIII are engaged in comparative experiments for two- and three-objective problems, respectively. The parameters of each meta-heuristic are set as follow:

\indent
Genetic Operation: The genetic operations used in the comparative algorithms MOEA/D, NSGA-II and NSGA-III are simulated binary crossover and polynomial mutation. According to \cite{deb2013evolutionary}, the crossover probability and the distribution index of the crossover operation are set to $p_c=1.0$ and $\eta_c=30$. Meanwhile, the mutation probability and the distribution index of the mutation operation are set to $p_m=1/n$ and $\eta_m=20$, where $n$ denotes the number of iterations. Moreover, the neighbourhood size of MOEA/D is set to 20.

\indent
Population Size: The population size of the comparative algorithms is set consistent with the number of reference vectors to ensure that all algorithms can search the objective space as fully as possible.

\indent
Termination Condition: According to \cite{li2020deep}, the maximum number of iterations of each comparative algorithm is set to 4000.

\section{Experimental results and discussions}
This section contains three parts. Section \uppercase\expandafter{\romannumeral4}.A shows the comparison and analysis of MOPN and DRL-MOA in terms of training time. Section \uppercase\expandafter{\romannumeral4}.B discusses the experimental results of the MOPN with other comparative methods on different testing problems. The results obtained from the comparative experiment are summarised in Section \uppercase\expandafter{\romannumeral4}.C.

\subsection{Comparison of training time}
TABLE \ref{tab:TAB3} shows the training time for models MOPN and DRL-MOA on all training problems. The first column lists the codes for the testing problems according to TABLE \ref{tab:TAB1}. The first row lists the training phases of MOPN and DRL-MOA, respectively, where MOPN-1 denotes the phase for training RM and MOPN-2 represents the phase for transfer learning. "Time/s", "Epochs" and "Sum" in the second row represent the average training time for each epoch, the number of training epochs and the total training time on each problem, respectively.
\begin{table*}[!htbp]
\centering
\caption{Comparative results using training time}
\setlength\tabcolsep{7pt}%调列距
\label{tab:TAB3}
    \begin{tabular}{llllllllll} \hline
    \multirow{2}[0]{*}{Problem} & \multicolumn{3}{c}{MOPN-1} & \multicolumn{3}{c}{MOPN-2} & \multicolumn{3}{c}{DRL-MOA} \\
    \cline{2-10}
          & Time/s & Epochs & Sum   & Time/s & Epochs & Sum   & Time/s & Epochs & Sum \\
    \hline
    T1O2  & 301.45 & 100   & 30145.34 & 1015.37 & 30    & 30461.14 & 331.12 & 500   & 165560.02 \\
    T2O2  & 270.29 & 100   & 27029.18 & 869.65 & 30    & 26089.57 & 313.36 & 500   & 156680.47 \\
    T2O3  & 402.91 & 100   & 40290.72 & 1417.58 & 30    & 42527.49 & 334.81 & 525   & 175775.25 \\
    \hline
    \end{tabular}
%\vspace{0.1in}
\end{table*}

\indent
According to columns 2 and 5 in TABLE \ref{tab:TAB3}, the training time per epoch of the MOPN is almost proportional to the problem scale of the training dataset. Rows 3 to 5 show that the training time of MOPN on problem T1O2 is slightly longer than that on problem T2O2. The reason is that, as an optimization objective, tour length is more complicated than altitude difference on T2O2. In addition, problem T2O3 has one more optimization objective than the other problems. Hence the training time of MOPN on this problem is longer than that of others. From column 10, it can be observed that the DRL-MOA has the similar training time on different problems. Since DRL-MOA requires training 5 epochs for each sub-model, the total number of training epochs is over 500. 

\indent
In MOPN, the training time using TS-RM for each problem is equal to the time shown in column 4 of TABLE \ref{tab:TAB3}. The training time using TS-TL is equal to the sum of time shown in columns 4 and 7 of TABLE \ref{tab:TAB3}. By comparing the total training time, it can be seen that the training costs of MOPN using TS-RM and TS-TL are approximately 20\% and 40\% of that of DRL-MOA, respectively.

\subsection{Comparison of optimization performance}
The following three parts are including in this subsection. 1) The experimental results of the proposed model and the comparative methods on the two-objective problems are specifically analysed. 2) The performance of all methods on the three-objective problem are discussed. In addition, the experimental results associated with parts 1) and 2) are obtained on the randomly generated testing datasets. 3) the evaluation indicators obtained from all methods on the generic RIns of TSPLIB are compared.

\subsubsection{Results on two-objective TSP}
\indent
 
\indent
TABLE \ref{tab:TAB4} gives the comparative results of MOPN-RM, MOPN-TL, DRL-MOA, NSGAII and MOEA/D on the testing datasets of two-objective TSP, where MOPN-RM and MOPN-TL represent the proposed models using TS-RM and TS-TL, respectively. The first column in TABLE \ref{tab:TAB4} lists the codes of testing datasets based on the MOTSP in TABLE \ref{tab:TAB1}. For example, T1O2S40 denotes the dataset corresponding to the T1O2 problem with problem scale of 40. "HV", "SPC", and "Time/s" in row 2 represent the indicators HV, SPC and Running time, respectively. Each DRL model was trained five times to generate five independent models. The average values of evaluation indicators obtained by these models are shown in TABLE \ref{tab:TAB4}. Similar to the above models, the comparative meta-heuristics are also executed five times on each RIns. In TABLE \ref{tab:TAB4}, the best results of each evaluation indicator on each testing dataset are shown in boldface.
\begin{table*}[!htbp]
\centering
\caption{Comparative results on two-objective TSP}
\label{tab:TAB4}
\setlength\tabcolsep{5pt}%调列距
        \begin{tabular}{llllllllllllllll} \hline
    \multirow{2}[0]{*}{\makecell{Problem\\Code}} & \multicolumn{3}{c}{MOPN-RM}  & \multicolumn{3}{c}{MOPN-TL} & \multicolumn{3}{c}{DRL-MOA} & \multicolumn{3}{c}{NSGAII} & \multicolumn{3}{c}{MOEA/D} \\
    \cline{2-16}
          & HV    & SPC   & Time/s & HV    & SPC   & Time/s & HV    & SPC   & Time/s & HV    & SPC   & Time/s & HV    & SPC   & Time/s \\ \hline
    T1O2S40 & \textbf{364.40 } & \textbf{0.64 } & \textbf{3.09 }  & 364.01  & 0.65  & 3.14  & 350.74  & 0.69  & 3.96  & 304.43  & 0.66  & 44.70  & 261.46  & 0.89  & 351.22  \\
    T1O2S70 & 1094.16  & \textbf{0.65 } & 5.97   & \textbf{1097.32 } & 0.65  & \textbf{5.95 } & 1035.60  & 0.67  & 6.87  & 807.94  & 0.67  & 53.01  & 710.20  & 0.88  & 368.21  \\
    T1O2S100 & 2086.98  & 0.64  & 9.37  & \textbf{2100.28 } & \textbf{0.64}  & \textbf{9.34 } & 1940.97  & 0.69  & 10.29  & 1366.14  & 0.72  & 62.78  & 1245.92  & 0.91  & 388.92  \\
    T1O2S150 & 5058.46  & 0.66  & 15.65  & \textbf{5109.48 } & \textbf{0.66}  & \textbf{15.31}  & 4678.31  & 0.71  & 16.20  & 3041.24  & 0.82  & 80.18  & 2858.57  & 0.94  & 419.84  \\
    T1O2S200 & 9171.28  & 0.69  & \textbf{18.99 } & \textbf{9288.68 } & \textbf{0.69}  & 19.02  & 8444.26  & 0.73  & 19.84  & 4887.04  & 0.89  & 87.09  & 4872.98  & 0.97  & 392.80  \\
    T2O2S40 & \textbf{216.61 } & 0.72  & \textbf{3.23 }  & 215.81  & \textbf{0.71}  & 3.30  & 216.37  & 0.77  & 4.11  & 185.39  & 0.73  & 44.89  & 152.82  & 0.95  & 366.56  \\
    T2O2S70 & 637.19  & \textbf{0.67 } & \textbf{5.70 } & \textbf{640.23 } & 0.68  & 5.82  & 630.92  & 0.76  & 6.61  & 468.45  & 0.69  & 49.48  & 400.81  & 0.90  & 354.50  \\
    T2O2S100 & 1457.76  & 0.71  & \textbf{10.20 } & \textbf{1470.03 } & \textbf{0.71}  & 10.23  & 1438.18  & 0.78  & 11.12  & 1021.98  & 0.76  & 65.73  & 922.71  & 0.92  & 420.14  \\
    T2O2S150 & 3036.88  & \textbf{0.68}  & \textbf{15.70 } & \textbf{3078.86 } & 0.70  & 15.74  & 2965.72  & 0.81  & 16.82  & 1835.10  & 0.85  & 78.77  & 1741.43  & 0.97  & 423.51  \\
    T2O2S200 & 5115.13  & \textbf{0.66 } & \textbf{20.43 } & \textbf{5200.67 } & 0.70  & 20.74  & 4961.84  & 0.83  & 21.59  & 2713.52  & 0.91  & 88.13  & 2696.16  & 0.99  & 419.33  \\
    \hline
    \end{tabular}
%\vspace{0.1in}
\end{table*}

\indent
It can be seen from columns 2, 5, 8, 11 and 14 in TABLE \ref{tab:TAB4} that the MOPN-TL obtains the best HV values on all testing datasets except on 40-city. The model that obtains the best HV value on 40-city dataset is MOPN-RM. This means that TS-TL can further improve the performance of MOPN based on RM. Nevertheless, MOPN-RM also has a competitive optimization capability compared to the other methods. According to columns 2, 5 and 8 in TABLE \ref{tab:TAB4}, the larger the problem scales of testing datasets, the more significant the advantage of the MOPN over DRL-MOA in terms of HV. This indicates that MOPN is more suitable for optimizing large-scaled MOCOPs than DRL-MOA. As a result, it can be concluded that the proposed model outperforms the other comparative methods in terms of optimization capability.

\indent
Based on columns 3, 6, 9, 12 and 15 in TABLE \ref{tab:TAB4}, it can be seen that the values of SPC of the MOPNs is better than these of all other methods on all RIns, that is, the Pareto front obtained by the MOPN is more uniform than others.

\indent
According to columns 4, 7, 10, 13 and 16 in TABLE \ref{tab:TAB4}, the running time of DRL models on all problems is much lower than those of the comparative meta-heuristics. Moreover, the difference in running time between these models is extremely small.

\begin{figure}
\centering
\includegraphics[width=1\linewidth]{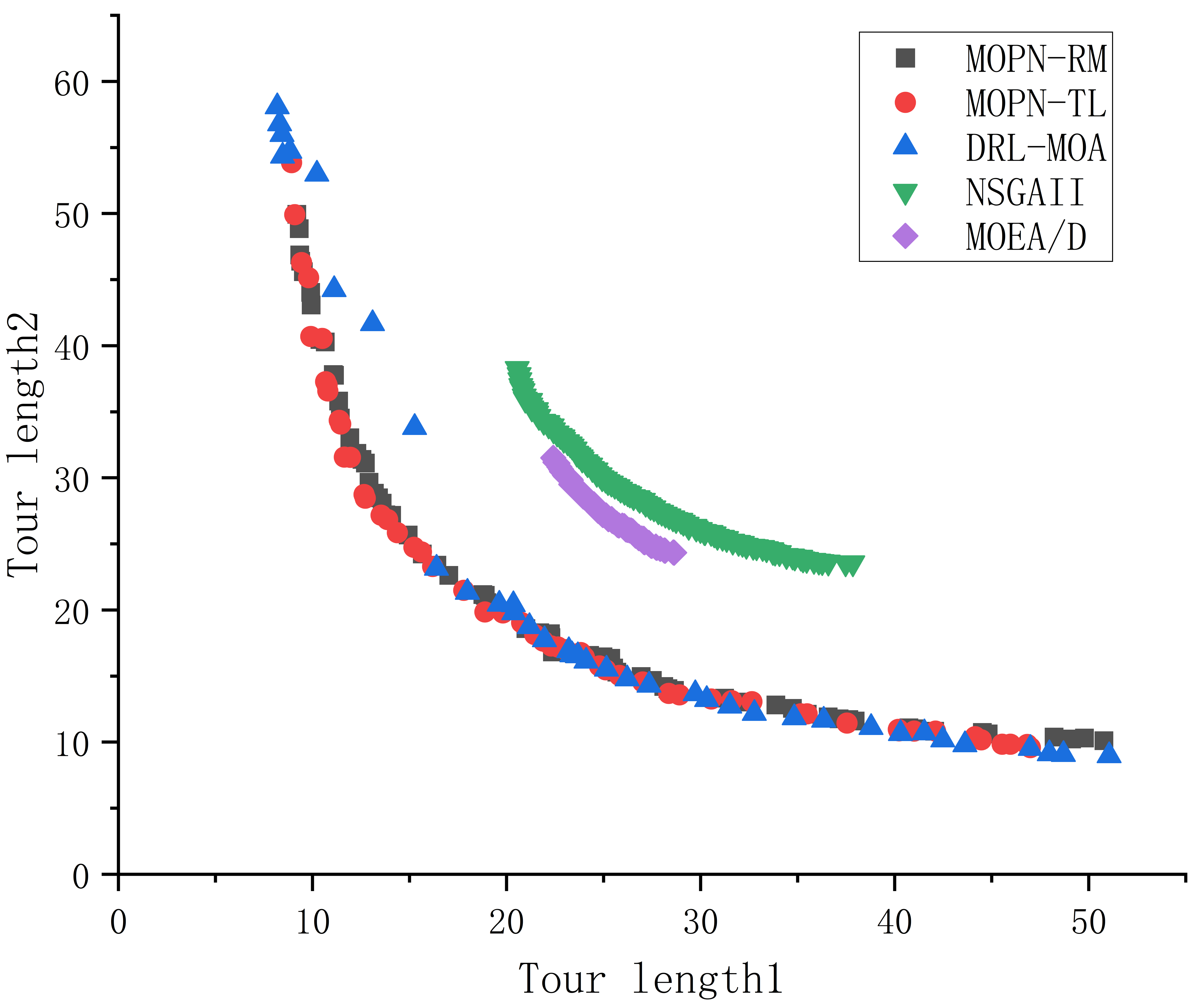}
\caption{Plots of non-dominated solutions on Ins. T1O2S100-1}
\label{fig:FIG4}
\end{figure}

\begin{figure}
\centering
\includegraphics[width=1\linewidth]{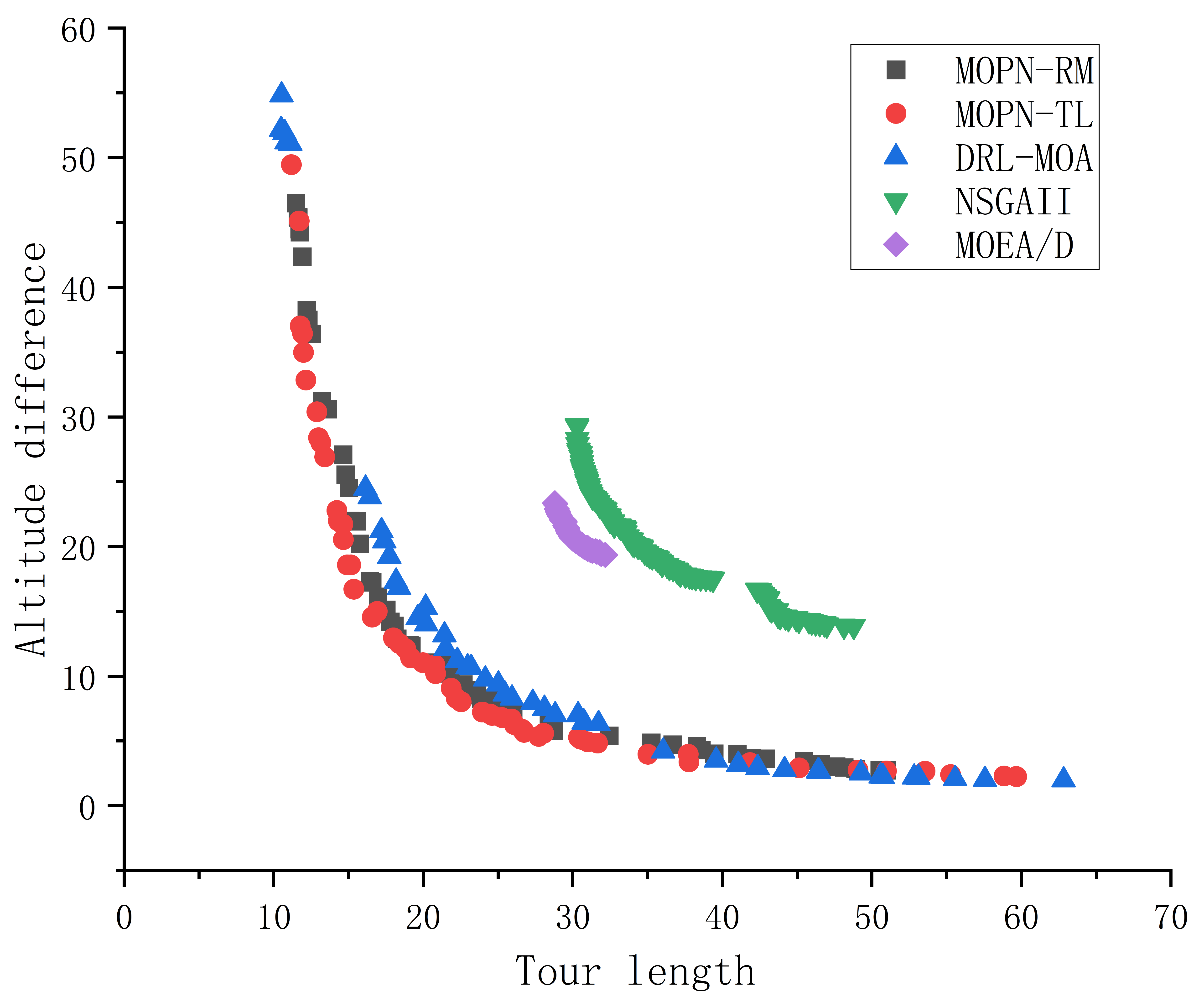}
\caption{Plots of non-dominated solutions on Ins. T2O2S150-1}
\label{fig:FIG5}
\end{figure}

\indent
Figs.\ref{fig:FIG4} and \ref{fig:FIG5} illustrate the distribution of non-dominated solutions to the RIns T1O2S100-1 and T2O2S150-1, which represent the first RIns of T1O2S100 and T2O2S150 testing datasets, respectively. Similar to the results in TABLE \ref{tab:TAB4}, the non-dominated solutions of each type of models in the two figures are all obtained through aggregating the results of the five independent models.

\indent
The X- and Y-axes in Fig.\ref{fig:FIG4} represent the two optimization objectives of the T1O2S100 problem. The closer the point to the axes, the higher the quality of the corresponding non-dominated solution on the relevant objectives. It can be observed from Fig.\ref{fig:FIG4} that the red points representing the solutions obtained by MOPN-TL are closer to the axis overall than the black points of MOPN-RM. Meanwhile, the two ends of the red point set are more dispersed than the set of black points, indicating that the proposed TS-TL improves the performance of MOPN and causes the Pareto front to be stretched towards both axes. The results of MOPN-RM and MOPN-TL shown in Fig.\ref{fig:FIG4} are consistent with those in TABLE \ref{tab:TAB4}. Moreover, the blue point set corresponding to DRL-MOA performs even slightly better than that of MOPN-TL in terms of Y-axis, but converges poorly in terms of X-axis, which can be attributed to the under-fitting in the training of the forward sub-models and the difficulty of transfer learning from the endpoints near Y-axis to the middle of the Pareto front. In addition, it can be seen from Fig.\ref{fig:FIG4} that the convergence and diversity of the comparative meta-heuristics are both poor.

\indent
The X- and Y-axes in Fig.\ref{fig:FIG5} represent the two optimization objectives of T2O2S150 problem. Similar to Fig.\ref{fig:FIG4}, the set of red points representing the solutions obtained by MOPN-TL in Fig.\ref{fig:FIG5} shows better convergence compared to MOPN-RM. At the same time, the Pareto front obtained by DRL-MOA has a large gap in the Y-axis direction.

\indent
It is worth mentioning in Figs.\ref{fig:FIG4} and \ref{fig:FIG5} that MOPN fails to find the better solutions at the ends of Pareto front than DRL-MOA. The reason is related to the running characteristic based on decomposition of DRL-MOA. When the objective vector of a sub-problem is parallel to the coordinate axes, it is equivalent for DRL-MOA to solve a single objective TSP using PN, which makes it easier to find the optimal value of the corresponding objective.

\subsubsection{Results on three-objective TSP}
\indent
 
\indent
TABLE \ref{tab:TAB5} presents the comparative results of MOPN-RM, MOPN-TL, DRL-MOA, NSGAIII and MOEA/D on the testing datasets of three-objective TSP. Similar to TABLE \ref{tab:TAB4}, MOPN-TL in TABLE \ref{tab:TAB5} obtains better HV values on the testing datasets with problem scales 70, 100, 150, and 200, respectively. It can be observed from columns 2, 5 and 8 that MOPN outperforms DRL-MOA in addressing the three-objective TSP. According to the results in TABLEs \ref{tab:TAB4} and \ref{tab:TAB5}, it can be seen that the optimization performance of MOPN on three-objective TSPs is more outstanding than that on two-objective TSPs. Meanwhile, the optimization effect of MOPN becomes more obvious as the problem scale increases. It can be found from columns 2, 3, 5 and 6 in TABLE \ref{tab:TAB5} that MOPN-TL obtains better HV values and worse SPC values than MOPN-RM on three-objective TSPs, that is, TS-TL makes the Pareto front obtained by MOPN more non-uniform while improving the convergence performance of MOPN. Moreover, from columns 4, 7, 10, 13 and 16 in TABLE \ref{tab:TAB5}, the running time of DRL models remains superior to that of the meta-heuristics on more complex three-objective problems.

\begin{table*}[!htbp]
\centering
\caption{Comparative results on three-objective TSP}
\label{tab:TAB5}
\setlength\tabcolsep{4pt}%调列距
    \begin{tabular}{llllllllllllllll} \hline
    \multirow{2}[0]{*}{\makecell{Problem\\Code}} & \multicolumn{3}{c}{MOPN-RM} & \multicolumn{3}{c}{MOPN-TL} & \multicolumn{3}{c}{DRL-MOA} & \multicolumn{3}{c}{NSGAIII} & \multicolumn{3}{c}{MOEA/D} \\
    \cline{2-16}
          & HV    & SPC   & Time/s & HV    & SPC   & Time/s & HV    & SPC   & Time/s & HV    & SPC   & Time/s & HV    & SPC   & Time/s \\
          \hline
    T2O3S40 & \textbf{5399.91 } & \textbf{0.69 } & \textbf{3.25 } & 5370.14  & 0.70  & 3.29  & 4173.56  & 0.95  & 4.14  & 3214.15  & 0.73  & 65.05  & 2699.60  & 0.82  & 392.17  \\
    T2O3S70 & 25461.34  & \textbf{0.69 } & \textbf{5.84 } & \textbf{25699.57 } & 0.69  & 5.87  & 17975.21  & 0.94  & 6.88  & 12568.10  & 0.76  & 70.82  & 10824.24  & 0.86  & 382.20  \\
    T2O3S100 & 60910.83  & \textbf{0.71 } & 9.26  & \textbf{62281.74 } & 0.72  & \textbf{9.24 } & 38310.55  & 0.98  & 10.36  & 24766.55  & 0.80  & 83.54  & 20777.49  & 0.86  & 410.18  \\
    T2O3S150 & 185864.41  & \textbf{0.72 } & \textbf{14.19}  & \textbf{192395.76 } & 0.74  & 14.20  & 102340.02  & 0.99  & 15.47  & 59225.99  & 0.85  & 96.91  & 47857.07  & 0.91  & 407.99  \\
    T2O3S200 & 474814.51  & \textbf{0.75 } & \textbf{19.18 } & \textbf{493633.24 } & 0.75  & 19.29  & 269076.38  & 1.00  & 20.73  & 137045.39  & 0.87  & 113.26  & 108606.35  & 0.91  & 422.88  \\
    \hline
    \end{tabular}%
%\vspace{0.1in}
\end{table*}
\begin{table*}[!htbp]
\centering
\caption{Comparative results on the generic RIns}
\label{tab:TAB6}
\setlength\tabcolsep{5pt}%调列距
    \begin{tabular}{llllllllllllllll} \hline
    \multirow{2}[0]{*}{\makecell{Problem\\Code}} & \multicolumn{3}{c}{MOPN-RM} & \multicolumn{3}{c}{MOPN-TL} & \multicolumn{3}{c}{DRL-MOA} & \multicolumn{3}{c}{NSGAII} & \multicolumn{3}{c}{MOEA/D} \\
    \cline{2-16}
          & HV    & SPC   & time  & HV    & SPC   & time  & HV    & SPC   & time  & HV    & SPC   & time  & HV    & SPC   & time \\
          \hline
    kroAB100 & 1879.16  & \textbf{0.65 } & \textbf{8.59 } & \textbf{1906.25 } & 0.66  & 8.77  & 1743.73  & 0.69  & 10.66  & 1209.98  & 0.73  & 59.88  & 1092.02  & 0.90  & 368.28  \\
    kroAB150 & 4880.51  & \textbf{0.66}  & \textbf{14.01 } & \textbf{4921.94 } & 0.70  & 14.14  & 4458.93  & 0.72  & 16.52  & 2787.70  & 0.83  & 81.94  & 2652.00  & 0.92  & 432.75  \\
    kroAB200 & 8887.19  & 0.71  & 21.97 & \textbf{9013.42 } & \textbf{0.67}  & \textbf{20.69 } & 8124.72  & 0.73  & 21.38  & 4494.26  & 0.88  & 98.57  & 4514.44  & 0.95  & 439.07  \\
    \hline
    \end{tabular}%
%\vspace{0.1in}
\end{table*}

\begin{figure}
\centering
\includegraphics[width=1\linewidth]{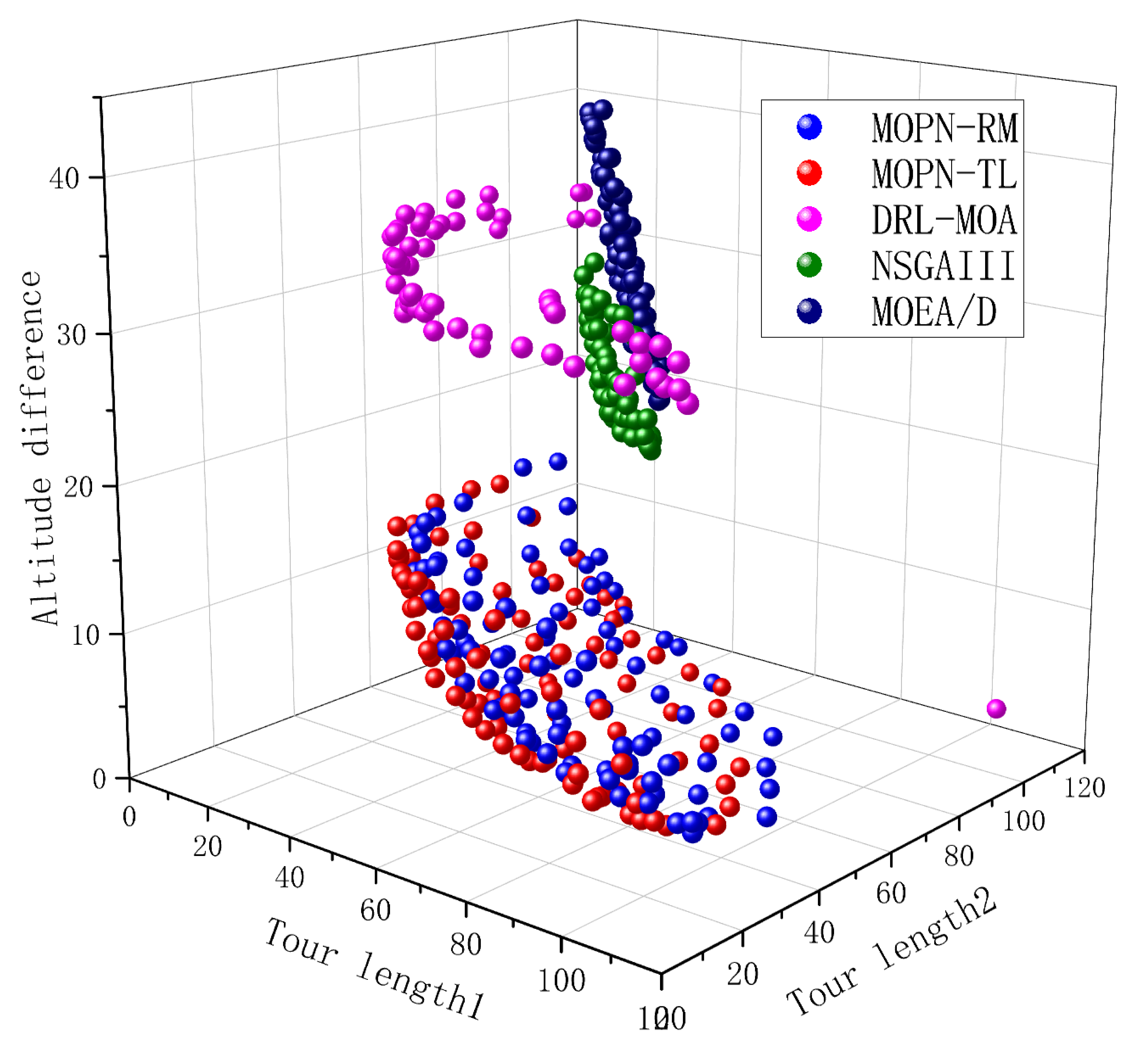}
\caption{Plots of non-dominated solutions on Ins. T2O3S200-1}
\label{fig:FIG6}
\end{figure}

\indent
Fig.\ref{fig:FIG6} shows the distribution of the non-dominated solutions to the RIns T2O3S200-1. The X-, Y-, and Z-axes represent the optimization objectives of the three-objective TSP, including two tour lengths and an altitude difference, respectively. According to Fig.\ref{fig:FIG6}, the convergence of the surface formed by the set of red points representing the solutions to MOPN-TL is better than those of MOPN-RM, indicating that the MOPN using TS-TL demonstrates better performance on addressing the three-objective TSP. The pink point set of DRL-MOA only displays a Pareto front for the objectives corresponding to the X- and Y-axes, indicating that DRL-MOA performs worse on optimizing the objective in terms of the z-axis. The single pink point in the lower right corner in Fig.\ref{fig:FIG6} is the solution obtained by the initial model of DRL-MOA. Similar to the case in Fig.\ref{fig:FIG5}, the Z-axis direction is relatively easy to be optimized. After the initial model falls into a local optimum in this direction, DRL-MOA is difficult to explore the other directions through transfer learning. From the two sets consisting of green points and purple points, respectively, corresponding to NSGAIII and MOEA/D in Fig.\ref{fig:FIG6}, there is a larger gap between the performance of these two meta-heuristics and that of MOPN-RM and MOPN-TL.

\subsubsection{Results on the generic RIns}
\indent
 
\indent
TABLE \ref{tab:TAB6} presents the comparative results of MOPN-RM, MOPN-TL, DRL-MOA, NSGAIII and MOEA/D on the generic RIns of T1O2. The first column in TABLE \ref{tab:TAB6} lists the codes of the three RIns, where the number at the end of each name represents the problem scale. In TABLE \ref{tab:TAB6}, the optimal HV values for the three RIns are obtained by MOPN-TL. The other results are also consistent with those in TABLE \ref{tab:TAB4}.

\subsection{Summary of the results}
From the above experimental results, it can be found that MOPN is able to solve MOTSP effectively. Different from the classical meta-heuristics, MOPN only needs to consume a little time on forward propagation to obtain a Pareto front of MOCOP. At the same time, MOPN has a stronger optimization capability than meta-heuristics.

\indent
Compared to the state-of-the-art model DRL-MOA, MOPN uses only one model to obtain the Pareto front for MOCOPs. Moreover MOPN shows better convergence and diversity than DRL-MOA on all types of testing datasets. The Pareto front obtained by MOPN is more homogeneous and without gaps. Additionally, MOPN-RM and MOPN-TL, compared to, DRL-MOA reduce the training time by about 80\% and 60\%, respectively.

\indent
By comparing MOPN-RM and MOPN-TL with each other, it can be seen that MOPN with TS-RM or TS-TL shows different advantages. With good optimization performance, MOPN-RM takes the shortest training time among all the comparative methods. Furthermore, MOPN-RM is insensitive to problem scales, so that a trained MOPN-RM model can be applied to problems with different scales. Therefore, MOPN-RM can be employed in the case where training time and generality of models are focused on. MOPN-TL can be employed in the other applications to obtain better optimization results than MOPN-RM.

\section{Conclusions}
MOCOPs, a type of classical mathematical optimization problems, widely appear in various real applications. Meta-heuristics have been applied to solve such problems for a long time. However, the iteration characteristic of meta-heuristics leads to a lot of time consumption during solving large-scaled MOCOPs, which makes meta-heuristics hard to be applied to real-time optimization.

\indent
Inspired by the recent work of DRL for single-objective optimization, this paper proposes a single-model MOPN to solve MOCOPs, whose running time is much less than that of meta-heuristics. Moreover, the training time of MOPN is also reduced by 60\% to 80\% compared to the state-of-the-art model DRL-MOA. Furthermore, the performance of the proposed model is significantly better than the comparative methods. The experimental results demonstrate that DRL has great potential and broad application prospects in addressing combinatorial optimization.

\indent
There are two valuable directions to be investigated in future research. On the one hand, other novel network architectures can be attempted to enhance model performance. To solve MOCOPs, this paper proposes a feasible single-model framework, where the networks of encoder, decoder and attention layer can be replaced by more efficient network architectures, such as graph convolutional network\cite{manchanda2019learning} and multi-head attention layer\cite{kool2018attention}. On the other hand, the optimization algorithms with DRL-based local search strategy\cite{chen2019learning,gao2020learn} can be taken into consideration to solve MOCOPs. Considering that DRL models do not require manual design, this type of algorithms is also appropriate for complex combinatorial optimization problems.

% or


%\appendices

%\section{}

%\section*{Acknowledgment}


\ifCLASSOPTIONcaptionsoff
  \newpage
\fi

\begin{IEEEbiography}[{\includegraphics[width=1in,height=1.25in,clip,keepaspectratio]{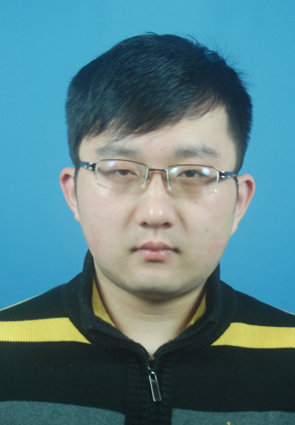}}]{Le-yang Gao}
received his master's degree in School of Computer Science and Technology from Anhui University, Hefei, P.R. China, in 2020, majoring in multi-objective optimization, is currently pursuing his Ph.D. degree in School of Computer Science and Technology, Anhui University, Hefei, P.R. China. His current research interests are deep reinforcement learning and multi-objective optimization.
\end{IEEEbiography}

\vspace{-2.0em}
\begin{IEEEbiography}[{\includegraphics[width=1in,height=1.25in,clip,keepaspectratio]{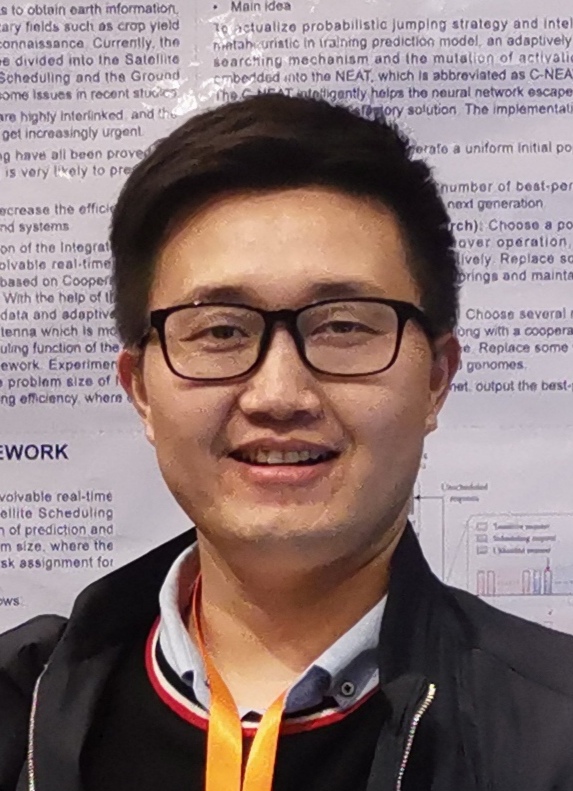}}]{Rui Wang}
received his Bachelor degree from the
National University of Defense Technology, P.R.
China in 2008, and the Doctor degree from the
University of Sheffield, U.K in 2013. Currently, he
is an Associate professor with the National University
of Defense Technology. His current research
interest includes evolutionary computation, multiobjective optimization and the development of algorithms applicable in practice. Dr. Wang received
the Operational Research Society Ph.D. Prize at
2016, and the National Science Fund for Outstanding
Young Scholars at 2021. He is also an Associate Editor for the IEEE Trans. on Evolutionary Computation, Swarm and Evolutionary Computation.
\end{IEEEbiography}

\vspace{-2.0em}
\begin{IEEEbiography}[{\includegraphics[width=1in,height=1.25in,clip,keepaspectratio]{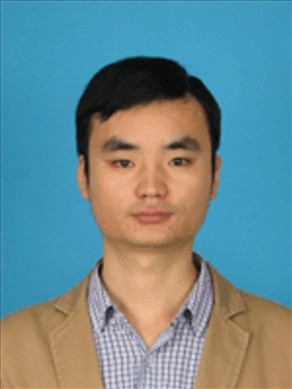}}]{Chuang Liu}
received his Ph.D. degree in Management Science and Engineering from University of Science and Technology of China, Hefei, Anhui, P.R. China, in 2021. His research interests include scheduling problems, routing problems, reinforcement learning, and evolutionary algorithms.
\end{IEEEbiography}

\vspace{-2.0em}
\begin{IEEEbiography}[{\includegraphics[width=1in,height=1.25in,clip,keepaspectratio]{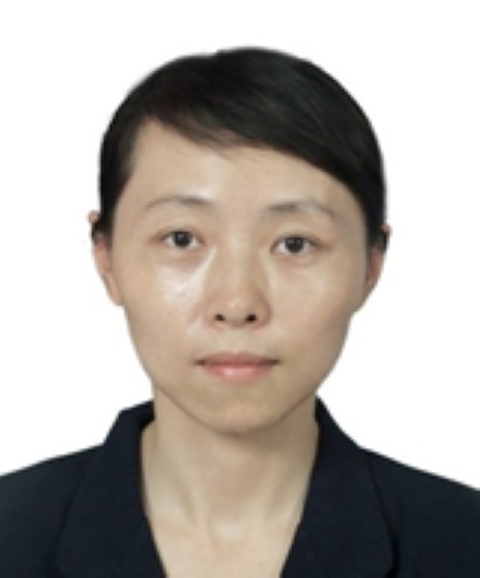}}]{Zhao-hong Jia}
received her Ph.D. degree in Management Science and Technology in 2008 from University of Science and Technology of China. She is a Professor with the School of Computer Science and Technology, Anhui University, Hefei. Her research interests include service computing, intelligent computation and its applications, service recommendation.
\end{IEEEbiography}

%\vfill


\end{document}